\pdfoutput=1 

\documentclass[sigconf, balance=false]{acmart}

\usepackage{enumitem}

\usepackage{multirow}

\usepackage{tikz-qtree}
\usepackage{pgfplots}
\pgfplotsset{compat=1.16}
\usetikzlibrary{shapes, positioning, arrows}

\newcommand\scalemath[2]{\scalebox{#1}{\mbox{\ensuremath{\displaystyle #2}}}}

\usepackage[linesnumbered]{algorithm2e}

\input{glyphtounicode}
\usepackage{xcolor}

\definecolor{tab20darkblue}{HTML}{4e79a7}
\definecolor{tab20darkgreen}{HTML}{59a14f}
\definecolor{tab20darkred}{HTML}{e15759}
\definecolor{tab20darkorange}{HTML}{f28e2b}
\definecolor{tab20darkturquoise}{HTML}{499894}
\definecolor{tab20darkgray}{HTML}{79706e}
\definecolor{tab20darkbrown}{HTML}{9d7660}
\definecolor{tab20darkpurple}{HTML}{b07aa1}
\definecolor{tabl20lighgreen}{HTML}{8cd17d}
\definecolor{tab20lightblue}{HTML}{a0cbd8}

\copyrightyear{2022}
\acmYear{2022}
\setcopyright{acmlicensed}\acmConference[WWW '22]{Proceedings of the ACM Web Conference 2022}{April 25--29, 2022}{Virtual Event, Lyon, France}
\acmBooktitle{Proceedings of the ACM Web Conference 2022 (WWW '22), April 25--29, 2022, Virtual Event, Lyon, France}
\acmPrice{15.00}
\acmDOI{10.1145/3485447.3511925}
\acmISBN{978-1-4503-9096-5/22/04}

\begin{document}
  
\title[EvoLearner: Learning Description Logics with Evolutionary Algorithms]{EvoLearner: Learning Description Logics with \texorpdfstring{\\}{} Evolutionary Algorithms}

\settopmatter{authorsperrow=1}

\author{%
Stefan~Heindorf,
Lukas~Blübaum,
Nick~Düsterhus,
Till~Werner,
Varun~Nandkumar~Golani,
Caglar~Demir,
Axel-Cyrille~Ngonga~Ngomo}
\affiliation{%
\mbox{\institution{DICE Group, Department of Computer Science, Paderborn University}\city{Paderborn}\country{Germany}}\\
\institution{\{heindorf, caglar.demir, axel.ngonga\}@upb.de,
\{lukasbl, nduester, wtill, vngolani\}@mail.upb.de}
}

\renewcommand{\shortauthors}{Heindorf et al.}



\begin{abstract}
Classifying nodes in knowledge graphs is an important task, e.g., for predicting missing types of entities, predicting which molecules cause cancer, or predicting which drugs are promising treatment candidates. While black-box models often achieve high predictive performance, they are only post-hoc and locally explainable and do not allow the learned model to be easily enriched with domain knowledge. Towards this end, learning description logic concepts from positive and negative examples has been proposed. However, learning such concepts often takes a long time and state-of-the-art approaches provide limited support for literal data values, although they are crucial for many applications. In this paper, we propose EvoLearner---an evolutionary approach to learn concepts in $\mathcal{ALCQ(D)}$, which is the attributive language with complement~($\mathcal{ALC}$) paired with qualified cardinality restrictions~($\mathcal{Q}$) and data properties~($\mathcal{D}$). We contribute a novel initialization method for the initial population: starting from positive examples, we perform biased random walks and translate them to description logic concepts. Moreover, we improve support for data properties by maximizing information gain when deciding where to split the data. We show that our approach significantly outperforms the state of the art on the benchmarking framework SML-Bench for structured machine learning. Our ablation study confirms that this is due to our novel initialization method and support for data properties.

\end{abstract}

\maketitle

\section{Introduction}%
\label{sec:introduction}

While deep learning has become popular over the last decade, its predictions are hardly explainable to humans, and attempts of peeking into the ``black box'' often focus on \emph{local} explainability rather than \emph{global} explainability, i.e., on single predictions rather than the model as a whole~\citep{Guidotti2019Survey}. However, when classifying nodes in knowledge graphs, e.g., predicting the type of entities, predicting which molecules cause cancer, or predicting which drugs are promising treatment candidates, it is often desirable to understand the model as a whole. This enables experts to double-check the model's plausibility and adapt the learned rules to reflect domain knowledge. Toward this end, interpretable models based on description logics have been proposed~\citep{Lehmann2009DL-Learner,Tran2017Parallel,Westphal2019SML-Bench}. Concepts in description logics can be directly mapped to OWL class expressions allowing to make use of the rich data ecosystem centered around the Web Ontology Language, OWL, the W3C standard at the core of the Semantic Web~\cite{Li2021Computing, Hogan2021Knowledge}. Tools for verbalizing OWL class expressions in natural language are readily available to explain the expressions to users~\citep{Moussallem2020NABU, Moussallem2020Generating, Ngonga2019Holistic}. OWL class expressions can be applied to web-based knowledge graphs such as DBpedia~\cite{Auer2007DBpedia}, Wikidata~\cite{Vrandecic2014Wikidata}, and YAGO~\cite{Suchanek2007Yago}, as well as to semantic data on websites, e.g., in the schema.org ontology~\cite{Hernich2015Schema.org} for type prediction~\cite[cf.][]{Paulheim2013Type, Zahera2021ASSET}, product classification~\cite{Zhang2019Product}, and discovery and querying of web APIs~\cite{Michel2019Enabling,Wolters2017Linking}. 

State-of-the-art approaches for concept learning are based on inductive logic programming with refinement operators~\citep{Fanizzi2018DLFoil,Lehmann2009DL-Learner,Lehmann2010Concept}: Starting from the most general concept $\top$ (also known as \texttt{Thing}), they refine the concept iteratively. For example, a concept is replaced by a subconcept or by multiple subconcepts joined via logical operators such as disjunction, conjunction or negation. In these approaches, the generation of candidate concepts is almost exclusively based on the ontology, and positive and negatives examples are only used to evaluate the generated concepts. In case of large ontologies, these approaches lead to a combinatorial explosion and many generated concepts never appear in the instance data leading to an unnecessarily large search space. For example, in the DBpedia ontology, a person can have over~100 different properties and ILP approaches would try all of them including combinations thereof. However, most of these properties (and combinations thereof) rarely occur in the instance data, e.g., it is seldom stated whether a person is left-handed or right-handed (object property ``handedness'') and neither is their hip size (data property ``hip size'').

Moreover, many knowledge bases contain large amounts of data properties connecting entities to literals, such as numeric values, which are crucial for good predictions. However, most state-of-the-art approaches for concept learning neglect data properties. Only \mbox{DL-Learner} \citep{Lehmann2010Concept} and \mbox{SPaCEL}~\citep{Tran2017Parallel} have some rudimentary support: DL-Learner divides the value range of a data property into bins containing approximately the same number of examples; \mbox{SPaCEL} considers all thresholds in a brute-force manner. Neither approach takes the distribution of positive and negative examples into account, resulting in suboptimal thresholds or long runtimes. DL-FOIL~\citep{Fanizzi2018DLFoil} does not support data properties at all.

In this work, we speed up and improve concept learning using a bottom-up approach, dubbed EvoLearner, that is based on random walks and evolutionary algorithms. Instead of refining the top concept $\top$, we start at the instance data by initializing the initial population of the evolutionary algorithm with biased random walks starting from the positive instances in the knowledge graph. We further refine the initial candidates by means of crossover and mutation operations. For data properties, we determine splits to maximize information gain in a way inspired by decision trees~\citep{Quinlan1986Induction}. Our approach significantly outperforms the state-of-the-art approaches CELOE~\cite{Lehmann2009DL-Learner} (top-down), Aleph~\cite{Muggleton1995Inverse} (bottom-up) and SPaCEL~\cite{Tran2012Approach} (hybrid) on 7 of 9 datasets available in the SML-Bench benchmarking framework~\citep{Westphal2019SML-Bench}. Our ablation study shows that both our random-walk initialization and support for data properties significantly contribute to our strong performance. Investigating predictive performance as a function of available runtime shows that EvoLearner outperforms CELOE and SPaCEL at all time points.

In the following, we briefly discuss related work in Section~\ref{sec:related-work} before introducing our EvoLearner in Section~\ref{sec:approach}. Section~\ref{sec:evaluation} evaluates our approach and Section~\ref{sec:discussion} discusses its strength and weaknesses. Section~\ref{sec:conclusion} concludes the paper.

\section{Related Work}%
\label{sec:related-work}

Most state-of-the-art approaches~\citep{Lehmann2009DL-Learner,Lehmann2010Concept,Tran2012Approach} for class expression learning employ inductive logic programming (ILP) with \emph{refinement operators}. The first ILP approaches GOLEM~\citep{Muggleton1990Efficient,Muggleton1991Inductive}, ProGOLEM~\citep{Muggleton2009ProGolem}, and Aleph~\citep{Muggleton1995Inverse} employed \emph{upward} refinement operators by means of least generalization. However, as \citet{Badea2000Refinement} argue, they generate overly specific concepts that tend to overfit. Hence, they propose tackling the problem by \emph{downward} refinement operators with subsequent approaches improving the downward refinement operators by means of heuristics~\citep{Lehmann2010Concept,Lehmann2011Class} and parallelization~\citep{Tran2012Approach}. The latest approaches, DL-FOIL~\citep{Fanizzi2018DLFoil} and SPaCEL~\citep{Tran2017Parallel}, employ a hybrid approach of \emph{upward} and \emph{downward} refinements. DL-FOIL, however, assumes the existence of a ``perfect'' concept covering exactly all positive and negative instances, which is hardly possible in realistic scenarios and causes the algorithm not to terminate. SPaCEL overcomes this problem by combining many partial descriptions. Our approach, Evo\-Learner, finds shorter concepts that are more likely to generalize, and in cases without a perfect solution, our approach terminates with a good approximation. In our evaluation, we show that  EvoLearner outperforms SPaCEL on most datasets.

While most ILP approaches employ refinement operators, there have been some attempts in the direction of \emph{evolutionary algorithms}. \mbox{\citet{Reiser1999Evolution}} showed that the ability of evolutionary approaches to search in a global space, allowing to escape local minima, can lead to an improvement over ILP algorithms. To our knowledge, \citet{Lehmann2007Hybrid} was the first to apply evolutionary algorithms to the task of concept learning in \emph{description logic}. He combined standard genetic programming~(GP) approaches with genetic refinement operators. For evaluation, he employs a single, small dataset comparing his genetic refinement approach to a standard GP approach. However, we were not able to reproduce his reported results, and in a pilot study, we found that our evolutionary approach works better with standard GP operators than with genetic refinements. \citet{Divina2006Evolutionary} gives an overview of evolutionary concept learning in first-order logic. In contrast, we specialize in description logics and contribute a sophisticated initialization strategy based on random walks~\citep{Fronczak2009Biased, Sinatra2011Maximal}.

\section{Evolutionary Concept Learning}%
\label{sec:approach}

In this section, we give a brief introduction to description logics and define the task of concept learning before introducing our novel approach EvoLearner for this task. We represent description logic concepts as abstract syntax trees which form the individuals of the evolutionary algorithms. The initial population of individuals is obtained via biased random walks originating from the positive examples. Subsequently, the population evolves from generation to generation by (1)~generating offspring via crossover operations, (2)~subjecting some of the offspring to mutations, and (3)~selecting the fittest individuals. Data properties are handled similarly to classical decision trees by maximizing information gain~\citep{Quinlan1986Induction}.

\makeatletter
\newcommand\setarraystretch[1]{%
  \noalign{\ifnum0=`}\fi
  \gdef\arraystretch{#1}%
  \global\setbox\@arstrutbox\hbox{%
    \vrule \@height\arraystretch\ht\strutbox%
    \@depth\arraystretch \dp\strutbox%
    \@width\z@}%
  \ifnum0=`{\fi}%
}
\makeatother

\begin{table}[tb]
  \centering
  \caption{Description logic constructs supported by Evo\-Learn\-er. For their semantics, we refer to Lehmann et al. \cite{Lehmann2010Concept}.}%
  \label{tab:description-logics}
  \footnotesize
  \setlength{\tabcolsep}{6pt}
  \renewcommand{\arraystretch}{1.15}
  \begin{tabular}{@{}ll@{}}
  \toprule
  \textbf{Syntax} & \textbf{Construct} \\
  \midrule
  $\mathcal{ALC}$ \\
  \midrule
  $r$           & abstract role           \\
  $b$           & Boolean concrete role   \\
  $d$           & numeric concrete role    \\[6pt]
  $\neg C$      & negation                \\
  $C \sqcup C$  & union                   \\
  $C \sqcap C$  & intersection            \\
  $\exists r.C$ & existential restriction \\
  $\forall r.C$ & universal restriction   \\
  \bottomrule
  \end{tabular}\hfill
  \begin{tabular}{@{}ll@{}}
  \toprule
  \textbf{Syntax} & \textbf{Construct} \\
  \midrule
  $\mathcal{Q}$ \\
  \setarraystretch{1.25}
  \midrule
  $\leq n\ r.C$ & max.\ cardinality restriction \\
  $\geq n\ r.C$ & min.\ cardinality restriction \\
  \midrule
  $\mathcal{(D)}$ \\
  \midrule
  $d \leq v $                              & max.\ numeric restriction ($v \in \mathbb{R}$) \\
  $d \geq v $                              & min.\ numeric restriction ($v \in \mathbb{R}$) \\
  $b = \mathit{true}$  & Boolean value restriction \\
  $b = \mathit{false}$ & Boolean value restriction \\
  \bottomrule
  \end{tabular}
\end{table}

\paragraph{Description Logics.}

Description logics~\cite{Kroetzsch2012Description} are widely used to express rules in knowledge bases, e.g., to express which instances belong to a class. These rules are represented by logical expressions such as the union $A \sqcup B$, intersection $A \sqcap B$, and negation $\neg A$ of concepts $A$ and $B$. While different description logics offer different trade-offs between expressiveness and reasoning complexity, in this paper, we employ the description logic $\mathcal{ALCQ(D)}$ because it contains the basic logical operators of all description logics ($\mathcal{ALC}$) as well as cardinality restrictions ($\mathcal{Q}$) and support for data properties ($\mathcal{D}$), which we found to be important for many real-world applications. Table~\ref{tab:description-logics} gives an overview of the different constructs.

\paragraph{Task: Concept Learning in Description Logics.}
We define the task of concept learning following \citet{Lehmann2010Concept}. Let $\mathcal{K}=(\mathcal{T}, \mathcal{A})$ be a knowledge base where $\mathcal{T}$ refers to the terminological box (TBox) expressed in $\mathcal{ALCQ(D)}$ and $\mathcal{A}$ refers to the assertion box (ABox). The ABox describes
conceptual facts $C(x)$ where $x \in N_I$ is an instance,%
\footnote{In this paper, we refer to description logic \emph{individuals} as \emph{instances} in order to distinguish them from evolutionary individuals which represent description logic concepts.}
relational facts $r(x, y)$ with $x,y \in N_I$,
Boolean facts $b(x, y)$ with $x \in N_I$ and $y \in \{\mathit{true}, \mathit{false}\}$,
and numeric facts $d(x, y)$ with $x \in N_I$ and $y \in \mathbb{R}.$
Given a knowledge base $\mathcal{K}$, positive examples $E^+\subseteq N_I$, and negative examples $E^- \subseteq N_I$, the objective is to find a concept $C$ in $\mathcal{ALCQ(D)}$ that covers as many of the positive examples as possible while covering as few of the negative examples as possible. In the ideal case, together with $\mathcal{K}' := \mathcal{K} \cup \{ C \}$, it should follow that $\mathcal{K}' \models C(e^+)$ for all $ e^+ \in E^+$ and $\mathcal{K'} \not\models C(e^-)$ for all $e^- \in E^-$, i.e., the concept $C$ provides an explanation for the positive examples but not for the negative examples. Note that it is not always possible to perfectly cover all positives and negatives in practice due to noisy training data with $E^{+} \cap E^{-}\neq \emptyset$. Moreover, the perfect solution $\{e_1^+\} \wedge  \ldots \wedge \{e_n^+ \} \wedge \neg \{e_1^-\}\wedge \ldots \wedge \neg \{e_m^-\}$ is often undesirable as it might not generalize to new, unseen data. Hence, the performance is measured with traditional metrics such as accuracy and $F_1$-measure~\citep{Westphal2019SML-Bench}.

\paragraph{Reasoning and Graph Representation of Knowledge Base.}
Following Lehmann~et~al.~\cite{Lehmann2010Concept,Lehmann2011Class}, we assume the knowledge base $\mathcal{K}$ to be static during concept learning and we employ a two-step reasoning process: (1)~We employ the OWL reasoner Pellet to derive the instances of named classes and the relationships between them. The resulting graph allows to (approximately) answer all instance checks.
(2)~Following a closed-world assumption, queries are evaluated by set operations between instances of named classes.

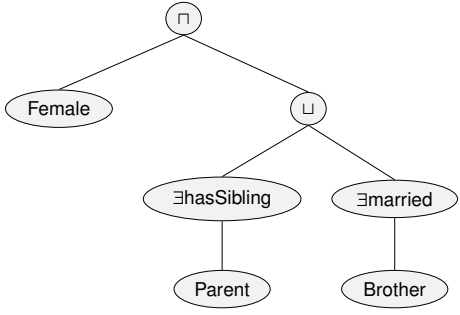
\begin{figure}[tb]
 	\centering
 	\tikzset{font=\fontfamily{phv}\selectfont}
	\begin{tikzpicture}[scale=0.8]
	\tikzset{every tree node/.style={minimum height=1.5em, draw, ellipse, fill=gray!10}, level distance=1.5cm,
    sibling distance=0.5cm}
	\Tree [.$\sqcap$ [.Female ]
				[.$\sqcup$ [.$\exists$hasSibling Parent ]
				[.$\exists$married [.Brother ] ] ] ]
	\end{tikzpicture}
	\caption{Tree representation of the concept \texttt{Female} $\sqcap$\\ (($\exists$\texttt{hasSibling.Parent}) $\sqcup$ ($\exists$\texttt{married.Brother})) from the Family dataset.}%
	\label{fig:tree-example}
	\Description{Tree representation of a concept from the Family dataset.}
\end{figure}

\paragraph{Tree Representation of Concepts.}
As is common in genetic programming, concepts are represented as trees: inner nodes represent operators and leaves represent terminals, which can be atomic concepts in case of object properties and numbers and Booleans in case of data properties. Figure~\ref{fig:tree-example} shows an example.

\subsection{Initialization via Biased Random Walks}%
\label{subsec:initialization}

We specifically tailored population initialization to concept learning to find good solutions as fast as possible, and we show that our method based on biased random walks significantly outperforms standard initialization methods for genetic pro\-gram\-ming, like \textsc{Grow}, \textsc{Full}, and \textsc{RampedHalfHalf}~\citep{Koza1992GP}. We seed the initial population of the evolutionary algorithm as follows: starting from a positive example in the knowledge graph, we perform a biased random walk and convert it to a concept which we add to the initial population. In this context, biased means that different outgoing edges are taken non-uniformly with a certain probability. Each generated concept consists of an atomic concept and sequences of role restrictions to describe its type and properties, respectively. Our algorithms are available as pseudocode in the appendix in Section~\ref{subsec:pseudocode}.

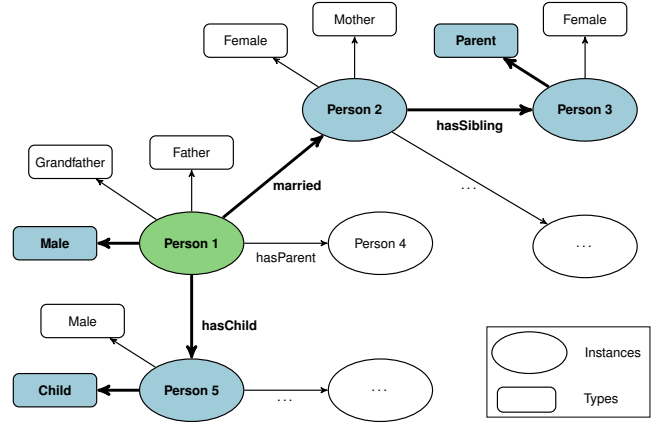
\begin{figure}[tb]
    \centering
    \tikzset{font=\fontfamily{phv}\selectfont}
    \resizebox{\columnwidth}{!}{%
    \begin{tikzpicture}
    \tikzstyle{instance} = [draw, minimum width=2.5cm, minimum height=1.5cm, ellipse, line width=0.8pt]
    \tikzstyle{type} = [draw, rounded corners, minimum width=2cm, minimum height=0.8cm, line width=0.8pt]
    \tikzstyle{arrow} = [->, thick,>=stealth']
    
    \tikzstyle{instanceLegend} = [draw, minimum width=1.8cm, minimum height=1cm, ellipse, line width=0.8pt]
    \tikzstyle{typeLegend} = [draw, rounded corners, minimum width=1.4cm, minimum height=0.6cm, line width=0.8pt]
    
    \matrix [draw,below left,row sep=0.3cm] at (11,-2) {%
      \node [instanceLegend] (p1){}; \node [right =0.3cm of p1] (p2){Instances}; \\
      \node [typeLegend] (p3){}; \node [right =0.55cm of p3] (p4){Types}; \\
    };
    
    \node[instance, fill=tabl20lighgreen] (I1) {\textbf{Person 1}};
    \node[type, left = of I1, fill=tab20lightblue] (T1) {\textbf{Male}};
    \node[type, above left = of I1] (T2) {Grandfather};
    \node[type, above = of I1] (T3) {Father};
    \draw[arrow, line width=2pt] (I1) -- (T1);
    \draw[arrow] (I1) -- (T2);
    \draw[arrow] (I1) -- (T3);
    
    \node[instance, above right = 3cm of I1, , fill=tab20lightblue] (I2) {\textbf{Person 2}};
    \draw[arrow, line width=2pt] (I1) to  node[right=0.1cm, pos=0.4]  {\textbf{married}} (I2);
    \node[type, above = of I2] (T4) {Mother};
    \node[type, above left =1cm of I2] (T5) {Female};
    \draw[arrow] (I2) -- (T4);
    \draw[arrow] (I2) -- (T5);
    
    \node[instance, right = 3cm of I2, , fill=tab20lightblue] (I3) {\textbf{Person 3}};
    \draw[arrow, line width=2pt] (I2) to  node[below=0.1cm, pos=0.5]  {\textbf{hasSibling}} (I3);
    \node[type, above = of I3] (T6) {Female};
    \node[type, above left =1cm of I3, fill=tab20lightblue] (T7) {\textbf{Parent}};
    \draw[arrow] (I3) -- (T6);
    \draw[arrow, line width=2pt] (I3) -- (T7);
    
    \node[instance, below = 1.75cm of I3] (I7) {\ldots};
    \draw[arrow] (I2) to  node[below=0.1cm, pos=0.5]  {\ldots} (I7);
    
    \node[instance, right = 2cm of I1] (I4) {Person 4};
    \draw[arrow] (I1) to  node[below=0.1cm, pos=0.5]  {hasParent} (I4);
    
    \node[instance, below = 2cm of I1, fill=tab20lightblue] (I5) {\textbf{Person 5}};
    \draw[arrow, line width=2pt] (I1) to  node[right=0.1cm, pos=0.6]  {\textbf{hasChild}} (I5);
    \node[type, left = of I5, fill=tab20lightblue] (T10) {\textbf{Child}};
    \node[type, above left = 1cm of I5] (T11) {Male};
    \draw[arrow, line width=2pt] (I5) -- (T10);
    \draw[arrow] (I5) -- (T11);
    
    \node[instance, right =2cm of I5] (I6) {\ldots};
    \draw[arrow] (I5) to  node[below=0.1cm, pos=0.5]  {\ldots} (I6);
    
    \end{tikzpicture}}
    \caption{Initialization of population: generating a concept via biased random walk (blue) originating from the chosen positive example (green).}
    \label{fig:init-example}
    \Description{Initialization of population: generating a concept via biased random walk originating from the chosen positive example.}
    \vspace{-1em}
\end{figure}

In the following, we describe our steps informally and illustrate them with the example shown in Figure~\ref{fig:init-example} from the Family dataset~\citep{Richards1995Automated}. Given positive and negative examples, the goal is to learn the concept of an \texttt{Uncle}:
\begin{equation}
\begin{split}
	\texttt{Male} \sqcap ((\exists \texttt{married.}\exists \texttt{hasSibling.Parent}) \ \sqcup \\ (\exists \texttt{hasSibling.Parent}))%
	\label{eq:expected-solution}
\end{split}
\end{equation}
The positive example we use to build an individual is \texttt{Person 1}.

\paragraph{Input and Precomputation}
Given a knowledge base~$\mathcal{K}$ and positive examples~$E^+$, we count all their types and super types: for each example $e \in E^+$, we increment the count $\mathit{ct}[t]$ for (super) type $t$ except for \texttt{Thing}.\footnote{In other words, $\mathit{ct}[C]$ expresses the number of instances of an atomic concept $C$.}

\begin{enumerate}
		\item\label{item1-rw} \textit{Select an example and one of its types}: Uniformly randomly pick an example \mbox{$e^{+} \in E^{+}$}. Then select one of its types~$t$, where each of its types is weighted according to its precomputed counts $\mathit{ct}[t]$. In a pilot study, we experimented with sampling the types uniformly randomly. However, weighting the types by their relative frequency performed slightly better. In our example, \texttt{Person 1} with its type \texttt{Male} is picked.
		
		\medskip
		\textbf{Example:} \texttt{Male}
		\medskip

		\item\label{item2-rw} \textit{Randomly select up to $\mathit{maxT}$ outgoing triples of~$e^{+}$, where $\mathit{maxT}$ is a hyperparameter}: Let $R = \{r \ | \ \mathcal{K} \models r(e^+, \cdot)\}$\ be the set of outgoing relations of $e^{+}$:%
\footnote{These are either object or data properties but not ``type relations'' since in description logics, the class of instances is expressed in the form of $C(x)$ rather than by means of relations $r$.}
		\begin{itemize}[noitemsep,topsep=0pt]
		    \item Uniformly randomly select up to $\mathit{maxT}$ relations $r$ from $R$ without replacement (duplicate relations are not considered to increase variety). For each selected relation $r$, uniformly randomly choose an object $o$, yielding the triple $r(e^{+}, o)$.
		    \item If less than $\mathit{maxT}$ triples $r(e^{+}, o)$ have been selected so far, uniformly randomly choose further triples $r(e^{+}, o)$ regardless of $r$ until either $\mathit{maxT}$ triples have been selected or all outgoing triples of~$e^{+}$ have been selected.
		\end{itemize}
		Having selected up to $\mathit{maxT}$ triples $r(e^{+}, o)$, their relations are added to the concept with either a union or an intersection, each with 50\% probability, as follows:
		\begin{itemize}[noitemsep,topsep=0pt]
		  \item In case $r$ is an object property, the relation is added as an existential restriction.
		  \item In case $r$ is a Boolean concrete role, the relation is added as a value restriction $r=o$, where $o \in \{\mathit{true}, \mathit{false}\}$ represents the actual value found in the graph
		  \item In case $r$ is a numeric concrete role, the split $\bar{v}_i$ closest to $o$ is selected from a precomputed set of potential splits (see Section~\ref{subsec:data_properties}). Then the value restriction $r \leq \bar{v}_i$ is added if $o \leq \bar{v}_i$ and the value restriction $r \geq \bar{v}_i$ otherwise.
		\end{itemize}
		We experimented with different values for $\mathit{maxT}$ and chose $\mathit{maxT}:=2$ because smaller values performed significantly worse in terms of $F_1$-measure, and larger values did not improve $F_1$ but tended to increase concept length (Table~\ref{tab:maxt}).
		In our example, we walked along the \texttt{married} relation to \texttt{Person~2} and along the \texttt{hasChild} yielding the following incomplete concept with placeholders ($\ldots$):
		
		\medskip
		
	    \textbf{Example:} $\texttt{Male} \sqcap (\exists \texttt{married} \ldots  \sqcap \exists \texttt{hasChild} \ldots)$
	    
	    \medskip
	    
	    \item\label{item3-rw} \textit{Complete incomplete subconcepts}.  For each incomplete subconcept $C$ with a placeholder, let $o$ be the entity in the knowledge graph corresponding to the placeholder.
	    \begin{itemize}[noitemsep,topsep=0pt]
	        \item With a 50\% chance, the subconcept $C$ is extended by another triple as in Step~2.
	        \item Finally, subconcept $C$ is completed by uniformly randomly selecting a type of $o$. If $o$ does not have a specific type, the top concept $\top$ (a.k.a.\ \texttt{Thing}) is selected.
	    \end{itemize}
	    
	    In our example, the \texttt{married}-subconcept is extended by another relation and a type, whereas the \texttt{hasChild}-subconcept is extended only by a type:
	    
		\medskip
		
		\textbf{Example:} $\texttt{Male} \sqcap ((\exists \texttt{married.}\exists \texttt{hasSibling.Parent})  \sqcap\\ (\exists \texttt{hasChild.Child}))$
\end{enumerate}

Steps 1--3 generate one concept, i.e., one evolutionary individual, and are repeated until the specified population size of the evolutionary algorithm is reached. Since the initialization is random, one positive example can serve as the starting point of multiple concepts. Our objective is not to find a perfect solution immediately, but to start with a diverse population providing good crossover material. As can be seen in the generated example solution, the part $\exists \texttt{hasChild.Child}$ is not part of the desired concept of an \texttt{Uncle}, see Solution~\eqref{eq:expected-solution}, whereas the second part is: $\exists \texttt{married.}\exists \texttt{hasSibling.Parent}$. Apart from starting at positive examples, we also experimented with starting at negative examples and negating the concept; however, this did not improve $F_1$-measure. The runtime of our method scales linearly with the population size, and in all our experiments with population sizes up to $1,000$ individuals, the initialization took less than two seconds.  Atomic concepts and relations not selected to be part of the initial population can be introduced later by our mutation operator, which deletes and inserts new concepts (see Section~\ref{subsec:evolutionary-algorithm}).

\subsection{Data Properties}%
\label{subsec:data_properties}

To overcome the problems of suboptimal data property thresholds and long runtimes outlined in Section~\ref{sec:introduction}, we adopt the idea of information gain~\citep{Quinlan1986Induction}, originally proposed for decision tree induction, and transfer it to concept learning: when generating value restrictions of the form $d \leq v$ and $d \geq v$, where $d$ is a numeric concrete role and $v$ is a threshold~(see Table~\ref{tab:description-logics}), we determine $v$ as to maximize information gain. Our evaluation results show that this approach significantly improves the performance.

Given a data property $d$ from the knowledge base $\mathcal{K}$ and a set $E$ of positive and negative examples, the split maximizing information gain is computed following~\citet{Quinlan1986Induction}: Let $V_d^{E} = \{v | \mathcal{K} \models d(e,v) \wedge e \in E \}$ be the set of all values for the data property~$d$ and examples~$E$. We sort all values $v_i \in V_d^E$, and for each pair of adjacent values $(v_i, v_{i+1})$, we compute the pair's mean $\bar{v_i} = (v_i + v_{i+1})/2$ for consideration as a potential split with information gain
\begin{equation}
\scalemath{0.91}{%
	\mathit{IG}(E, \bar{v_i}) = H(E) - H(E|\bar{v_i})
   =  H(E) - \left(\frac{|E_L|}{|E|} H(E_L) + \frac{|E_R|}{|E|}  H(E_R)\right)\,,}
\end{equation}
where $E_i^L = \{e| \left(\mathcal{K} \models d(e, u)\right) \wedge e \in E \wedge u \leq \bar{v_i} \}$ and $E_i^R = \{e|\left(\mathcal{K} \models d(e, u)\right) \wedge e \in E \wedge u > \bar{v_i} \}$ are the two sets produced by splitting $E$ on $\bar{v_i}$. In the formula, $H$ refers to the entropy
\begin{equation}
\scalemath{0.91}{%
		H(E) = - \sum_{c \in \{+, -\}} \Pr(E = c) \log \Pr(E = c)\,,}
\end{equation}
with $\Pr(E=c)$ denoting the fraction of positive/negatives examples in $E$; for example, with 4 positive and 3 negative examples, $H(E) = -4/7 \log(4/7) - 3/7 \log(3/7)$.
As proven by \citet{Fayyad1992Handling}, if all examples with values $v_i$ and $v_{i+1}$ belong to the same class, a split between the two values cannot lead to a partition maximizing information gain. Hence, at most $\min(|E^{+}|, |E^{-}|) < |E|$ splits need to be tried, yielding a runtime of $O(|E| \cdot \log |E|)$ due to the sorting.

When computing the graph representation of the knowledge base (see beginning of Section~\ref{sec:approach}), we precompute up to $k$ splits $\bar{v_i}$ per data property. We do so in a greedy fashion and take interactions between data properties into account as follows: We start with the set $E=E^{-} \cup E^{+}$ of positive and negative examples, and for each data property $d$ we compute the split $\bar{v}_i$ maximizing information gain along with its sets $E_i^L$ and $E_i^R$ which we add to a list of example sets. Once we have computed the best split for each data property, we sort the list of example sets according to entropy in descending order. Then for each data property $d$ and for each example set $E$ in the list, we compute the split that maximizes information gain. The process terminates once we have found $k$ splits per data property or there are less than $k$ splits per property. Intuitively, our computation is equivalent to computing one decision tree for each data property and traversing these decision trees in a breadth-first search from tree level to tree level, where each level is traversed in the order of decreasing entropy until we have found $k$ splits per data property. In our experiments, we chose $k=10$, and the whole computation of splits took less than a second for all learning problems. In future work, further non-precomputed splits might be obtained on the fly.

Within our approach, we add the precomputed splits to our graph representation of the knowledge base and employ them at two points: (1)~during the initialization phase described in Section~\ref{subsec:initialization}, when a numeric property is selected, we select the \emph{closest} split to the actual value of the data property; (2)~during the mutation described in Section~\ref{subsec:evolutionary-algorithm}, when a new subtree is randomly generated, a \emph{random} split from the precomputed splits is selected. For Boolean value restrictions, we pick the actual data value in the initialization and the mutation operator randomly picks a value $\mathit{true}$ or $\mathit{false}$; for cardinality restrictions, we pick a value $\{1, 2, \ldots, N\}$ where $N$ is a hyperparameter, which was set to $5$ in our experiments. In future work, this hyperparameter might be determined automatically based on the underlying data and it might vary from property to property.

\subsection{Selection, Crossover, and Mutation}%
\label{subsec:evolutionary-algorithm}

Starting from an initial population of evolutionary individuals represented as concept trees, selection, crossover, and mutation operators randomly generate new offspring based on their parents (algorithm \texttt{eaSimple}\cite{back2018evolutionary}).

To select the fittest individuals, a tournament selection is performed in a first pass over the population (\texttt{selTournament}): Let $n$ be the population size. Then $n$ tournaments are performed, where in each tournament a small number of individuals $k$ is chosen uniformly randomly with replacement and the fittest individual among them is selected. In a second pass over the population,  a crossover operator combines the information of two parents generating two descendants. In a third pass, a mutation operator randomly modifies single individuals. In our experiments, we employ one-point crossover (\texttt{cxOnePoint}) and the uniform mutation operation (\texttt{mutUniform}).%
\footnote{\url{https://deap.readthedocs.io/en/master/api/tools.html}}
The one-point crossover operator randomly chooses a cut point (i.e., a node in the parse tree) in both individuals and swaps the subtrees below the cut points. The mutation operation uniformly randomly selects a node in the tree and replaces it with a randomly generated tree: the tree is generated with the \textsc{RampedHalfHalf} method with height 1 to 3~\citep{Koza1992GP}. Following~\citet{Koza1992GP}, we employ a tournament size of~$k=7$ and apply the crossover operation to two consecutive individuals in a population with 90\% probability and the mutation operation with 10\% probability. In a pilot study, we experimented with further crossover operators (e.g., \texttt{cx\-One\-Point\-Leaf\-Biased}), mutation operators (e.g., \texttt{mut\-Shrink}, \texttt{mut\-Node\-Re\-place\-ment}, \texttt{mut\-Insert},  Table~\ref{tab:mutation}), and probabilities. Many combinations yielded good results and our chosen combination only slightly outperformed other combinations we tried. The whole process is called a generation and is repeated until the maximum number of generations is reached or a timeout occurs. The population size stays constant.

\subsection{Fitness Function and Bloat Control}

We measure the fitness of each evolutionary individual, i.e., each concept~$C$, in terms of $\mathit{accuracy}$ on the positive and negative examples in the training set. $\mathit{Accuracy}$ is defined as the ratio of correct predictions among all predictions
\begin{equation}
\mathit{accuracy}(C) = \frac{| E^{+} \cap R(C)| + |E^{-} \setminus R(C)|}{|E^{+}| + |E^{-}|}\,,
\end{equation}
with $R(C)$ denoting the individuals retrieved for concept $C$ via our graph representation. To control bloat, i.e., overly long concepts that do not generalize to new, unseen data, we apply linear parametric parsimony pressure~\citep{Luke2006Bloat} yielding the fitness function
\begin{equation}
		\mathit{fitness}(C) := \mathit{accuracy}(C) \cdot x - \mathit{length}(C)\,,
\end{equation}
where the $\mathit{length}(C)$ of an individual denotes the number of nodes in the tree (e.g., length~7 in Figure~\ref{fig:tree-example}) and where $x$ is a weighting factor that can be set as a hyperparameter. In a pilot study, as per~\citet{Luke2006Bloat}, we experimented with different values for~$x$ between 1,024 and 8,192 and decided to use $x=2,048$. We obtained similar results for a wide range of values. However, smaller values slightly decreased performance in terms of $F_1$-measure. Larger values were only beneficial for few datasets and learning problems.

In addition, we set a static depth limit $d$ for the tree depth of individuals since bloat control often performs better if combined with such a limit~\citep{Luke2006Bloat}. In our experiments, we set $d=17$ as previously done in the literature~\citep{Luke2006Bloat, Koza1992GP}. We also experimented with other bloat control methods such as proportional and double tournament~\citep{Luke2006Bloat} without significant improvements.

\section{Evaluation}%
\label{sec:evaluation}

After introducing our evaluation setup in Section~\ref{subsec:evaluation-setup}, we compare our approach with state-of-the-art approaches in Section~\ref{subsec:evaluation-best-approach}. In Sections~\ref{subsec:evaluation-ablation-analysis} and~\ref{subsec:evaluation-walk-variants}, we perform an ablation analysis to identify the components that contribute most towards our high performance. Section~\ref{subsec:evaluation-dataset-characteristics} studies the performance based on dataset characteristics. How the learned concepts evolve from generation to generation is exemplified in Section~\ref{subsec:evaluation-example}, and Section~\ref{subsec:effectiveness-over-runtime} investigates performance as a function of runtime.

\subsection{Evaluation Setup}%
\label{subsec:evaluation-setup}

Our evaluation was performed with SML-Bench~\citep{Westphal2019SML-Bench}, a benchmarking framework for structured machine learning.

\paragraph{Datasets and Learning Problems}

\begin{table}[tb]
\centering
\caption{Overview of the datasets in terms of number of instances ($N_I$), axioms, atomic concepts, properties, expressiveness and positive and negative examples ($E^+$, $E^-$). The statistics were obtained with Prot\'eg\'e.}%
\label{tab:evaluation-datasets}
\setlength{\tabcolsep}{1pt}
\renewcommand{\arraystretch}{1.1}
\footnotesize
\begin{tabular}{@{}l@{\hskip -7pt}rrrrrrrr@{}}
  \toprule
   & \textbf{Instances} & \textbf{Axioms} & {\textbf{Atomic}} & \textbf{Object} & \textbf{Data} & \textbf{Expres-} & $|E^{+}|$ & $|E^{-}|$  \\
  \textbf{Dataset} & & & {\textbf{Concepts}} & \textbf{Prop.} & \textbf{Prop.} & \textbf{siveness} & &  \\
  \midrule
  Carcinogenesis & 22,372 &    74,566 & 142 &  4 &  15 & $ \mathcal{ALC(D)}$  & 162 & 136 \\
  Family         &    202 &     1,829 &  18 &  4 &   0 & $ \mathcal{ALC}$     &  38 &  38 \\
  Hepatitis      &  6,812 &    79,935 &  14 &  5 &  12 & $ \mathcal{ALE(D)}$  & 206 & 294 \\
  Lymphography   &    148 &     2,193 &  49 &  0 &   0 & $ \mathcal{AL}$      &  81 &  67 \\
  Mammographic   &    975 &     6,808 &  19 &  3 &   2 & $ \mathcal{AL(D)}$   & 445 & 516 \\
  Mutagenesis    & 14,145 &    62,066 &  86 &  5 &   6 & $ \mathcal{AL(D)}$   &  13 &  29 \\
  NCTRER         & 10,209 &   103,070 &  37 &  9 &  50 & $ \mathcal{ALCI(D)}$ & 131 &  93 \\
  Premier League & 11,319 & 2,153,818 &  9 & 13 & 202 & $ \mathcal{ALEH(D)}$ &  40 &  41 \\
  Pyrimidine     &     74 &     2,080 &   1 &  0 &  27 & $ \mathcal{AL(D)}$   &  20 &  20 \\
  \bottomrule
\end{tabular}
\end{table}

Table~\ref{tab:evaluation-datasets} gives an overview of the nine datasets we employ for our evaluation. In contrast to SML-Bench's original configuration~\citep{Westphal2019SML-Bench},
we added the Family dataset~\citep{Lehmann2009DL-Learner}, which is provided as part of DL-Learner and often used to evaluate structured machine learning approaches~\citep{Tran2017Parallel}.
Per dataset, the SML-Bench framework provides one learning problem, i.e., one set of positive and negative examples. For the Family dataset, the concept of an \texttt{Uncle} is to be learned.

\paragraph{Baselines}

We compare our novel approach EvoLearner with four baselines: DL-Learner (CELOE), DL-Learner (OCEL), Aleph, and SPaCEL\@. Except for SPaCEL, all approaches are part of the SML-Bench framework~\citep{Westphal2019SML-Bench}. CELOE and OCEL employ inductive logic programming with downward refinement operators with different heuristics. Aleph serves as an example of a bottom-up approach~\citep{Muggleton1995Inverse}. SPaCEL~\citep{Tran2017Parallel} employs a hybrid approach of upward and downward refinements. We also experimented with DL-FOIL~\citep{Fanizzi2008DL-FOIL,Fanizzi2018DLFoil,Rizzo2020DL-Foil}; however, it assumes the existence of a perfect solution, which covers all positive examples and none of the negative examples. Since almost none of the SML-Bench learning problems have such a perfect solution with a reasonable length, DL-FOIL did not terminate and we excluded it from our experiments. Additionally, we experimented with Metagol~\citep{Muggleton2013MetainterpretiveLA}. However, Metagol did not terminate within our timeout for any of the datasets, corroborating previous findings that it has problems with larger datasets and that finding a suitable set of metarules for each dataset is a challenging, unsolved task~\citep{Dumancic2020Comparative,Cropper2020Inductive}. Hence, we excluded it from our experiments.

\paragraph{Hyperparameters}
As outlined in Section~\ref{sec:approach}, we employ a population size of~800 with~200 generations, a tournament size of~7, a crossover probability of~90\%, a mutation probability of 10\%, and a static depth limit of 17.

\paragraph{Cross-validation and Evaluation Metrics}
As SML-Bench~\citep{Westphal2019SML-Bench}, we employ 10-fold cross-validation, i.e., all positive and negative examples for each learning problem are randomly shuffled and split into 10~folds. In each round, the examples of one fold are used as the validation set and the other folds are used as the training set. Finally, the performance on all validation folds is averaged. As commonly done~\citep{Westphal2019SML-Bench}, we evaluate the predictive performance of the approaches in terms of accuracy and $F_1$-measure, and the explainability in terms of concept length~\citep{Tran2017Parallel}. Following SML-Bench, we employ a timeout of 5~minutes per fold.

\paragraph{Reproducibility}
To ensure the reproducibility of our experiments, the code underlying our research is publicly available.%
\footnote{\url{https://github.com/dice-group/EvoLearner}}
It enables those who wish to follow up on our work to replicate the plots and performance values reported.
We implemented EvoLearner in Python~3.6.9 with the evolutionary framework DEAP~\citep{Fortin2012DEAP}~1.3.1  and the ontology framework Owlready2~\citep{Lamy2017OWL}~0.29. For evaluation, we employ SML-Bench~\citep{Westphal2019SML-Bench}~0.3.0, DL-Learner~\citep{Lehmann2009DL-Learner}~1.4.0, and the latest version of SPaCEL~\citep{Tran2017Parallel}. Our experiments were run on a machine with 32~GB memory and an Intel Core i7-9750H with 2.6~GHz.

\subsection{Evaluation of EvoLearner}%
\label{subsec:evaluation-best-approach}

\begin{table}[tbp]
  \centering
  \caption{Evaluation results of EvoLearner and four state-of-the-art baselines in terms of accuracy, $\boldsymbol{F_1}$-measure, and concept length obtained via 10-fold cross-validation on 9~learning problems with a maximum execution time of 5~minutes per fold. In addition to the mean value, we report the folds' standard deviation. EvoLearner is initialized via random walks and supports data properties. The results of Evo\-Learner and SPaCEL are averaged over 3 runs since they contain randomness.}%
  \label{tab:evaluation-results}
  \footnotesize
  \setlength{\tabcolsep}{1.5pt}
  \renewcommand{\arraystretch}{1.08}
  \begin{tabular}{@{}lccccc@{}}
    \toprule
    \multicolumn{6}{c}{\textbf{Accuracy}}\\
    \midrule
     & \textbf{EvoLearner} & \textbf{DL-Learner} & \textbf{DL-Learner} & \textbf{Aleph} & \textbf{SPaCEL}  \\
    \textbf{Learn.~Problem} & \textbf{(ours)}  & \textbf{(CELOE)} & \textbf{(OCEL)} &  &  \\
    \midrule
    Carcinogenesis &  \textbf{0.64} $\boldsymbol{\pm}$ \textbf{0.16}   & $0.55 \pm 0.02$ & \textit{no results} & $0.48 \pm 0.10$ & $0.51 \pm 0.10$ \\
    Family & 1.00 $\pm$ 0.01 & $0.97 \pm 0.05$ & \textbf{1.00} $\boldsymbol{\pm}$ \textbf{0.00} & --- & $0.97 \pm 0.08$  \\
    Hepatitis & \textbf{0.82} $\boldsymbol{\pm}$ \textbf{0.06}  & $0.49 \pm 0.06$ & \textit{no results} & $0.67 \pm 0.05$ & \textit{no results}  \\
    Lymphography & 0.81 $\pm$ 0.12  & $0.70 \pm 0.15$ & \textbf{0.85} $\boldsymbol{\pm} $ \textbf{0.09} & 0.83 $\pm$ 0.10 & $0.71 \pm 0.14$  \\
    Mammographic & \textbf{0.83} $\boldsymbol{\pm}$ \textbf{0.05} & $0.49 \pm 0.02$ & $0.82 \pm 0.04$ & $0.65 \pm 0.04$ &  $0.70 \pm 0.05$ \\
    Mutagenesis & \textbf{1.00} $\boldsymbol{\pm} $ \textbf{0.00}  & $0.94 \pm 0.13$ & \textit{timeout} & $0.72 \pm 0.25$ & \textbf{1.00} $\boldsymbol{\pm} $ \textbf{0.00} \\ 
    NCTRER & \textbf{1.00} $\boldsymbol{\pm}$ \textbf{0.00}&  $0.59 \pm 0.03$ & $0.94 \pm 0.06$ & $0.72 \pm 0.14$ &  \textbf{1.00} $\boldsymbol{\pm}$ \textbf{0.00}  \\
    Premier League & \textbf{1.00} $\boldsymbol{\pm}$ \textbf{0.00} & $0.99 \pm 0.04$ & $0.85 \pm 0.01$ & 0.95 $\pm $ 0.09 & $0.98 \pm 0.04$ \\
    Pyrimidine & $0.90 \pm 0.15$ & $0.82 \pm 0.17$ & $0.85 \pm 0.24$ & \textbf{0.95} $\boldsymbol{\pm} $ \textbf{0.16} & $0.87 \pm 0.24$ \\
    \bottomrule
  \end{tabular}
  \setlength{\tabcolsep}{1.6pt}
  \begin{tabular}{@{}lccccc@{}}
    \toprule
    \multicolumn{6}{c}{\textbf{$\boldsymbol{F_1}$-measure}}\\
    \midrule
     & \textbf{EvoLearner} & \textbf{DL-Learner} & \textbf{DL-Learner} & \textbf{Aleph} & \textbf{SPaCEL} \\
    \textbf{Learn.~Problem} & \textbf{(ours)} & \textbf{(CELOE)} & \textbf{(OCEL)} &  & \\
    \midrule
    Carcinogenesis &  $0.70 \pm 0.12$ &  \textbf{0.71} $\boldsymbol{\pm}$ \textbf{0.01} & \textit{no results} & $0.46 \pm 0.12$ & $0.60 \pm 0.08$  \\
    Family & 1.00 $\pm$ 0.01  &  $0.98 \pm 0.05$ & \textbf{1.00} $\boldsymbol{\pm}$ \textbf{0.00} & --- & $0.97 \pm 0.11$ \\
    Hepatitis & \textbf{0.79} $\boldsymbol{\pm}$ \textbf{0.08} & $0.61 \pm 0.03$ & \textit{no results} & $0.38 \pm 0.12$ & \textit{no results}  \\
    Lymphography & 0.84 $\pm$ 0.09 &  $0.78 \pm 0.10$ & \textbf{0.85} $\boldsymbol{\pm}$ \textbf{0.10}  & $0.84 \pm 0.09$ & $0.75 \pm 0.13$  \\
    Mammographic & \textbf{0.81} $\boldsymbol{\pm}$ \textbf{0.06}&  $0.64 \pm 0.01$ & $0.78 \pm 0.08$ & $0.48 \pm 0.08$ & $0.64 \pm 0.06$ \\
    Mutagenesis & \textbf{1.00} $\boldsymbol{\pm} $ \textbf{0.00} &  $0.93 \pm 0.14$ & \textit{timeout} & $0.43 \pm 0.47$ & \textbf{1.00} $\boldsymbol{\pm} $ \textbf{0.00} \\
    NCTRER & \textbf{1.00} $\boldsymbol{\pm}$ \textbf{0.00}&  $0.74 \pm 0.01$ & $0.94 \pm 0.06$ & $0.71 \pm 0.18$ & \textbf{1.00} $\boldsymbol{\pm}$ \textbf{0.00} \\
    Premier League & \textbf{1.00} $\boldsymbol{\pm}$ \textbf{0.00} & $0.99 \pm 0.04$ & $0.81 \pm 0.13$ & 0.94 $\pm $ 0.11 & $0.98 \pm 0.04$ \\
    Pyrimidine & \textbf{0.91} $\boldsymbol{\pm} $ \textbf{0.14}  &  $0.84 \pm 0.15$ & $0.84 \pm 0.22$ & $0.90 \pm 0.32$  & $0.86 \pm 0.29$  \\
    \bottomrule
  \end{tabular}
  \setlength{\tabcolsep}{4.5pt}
  \begin{tabular}{@{}lcccc@{}}
    \toprule
    \multicolumn{5}{c}{\textbf{Concept Length}} \\
    \midrule
    & \textbf{EvoLearner} & \textbf{DL-Learner} & \textbf{DL-Learner} & \textbf{SPaCEL} \\
    \textbf{Learn.~Problem} & \textbf{(ours)}  & \textbf{(CELOE)} & \textbf{(OCEL)} \\
    \midrule
    Carcinogenesis &  $23.47 \pm 4.10$ &  $\phantom{0}3.90 \pm 0.32$ & \textit{no results} & $1093.30 \pm 82.39$ \\
    Family & $10.87 \pm 1.90$  &  $\phantom{0}9.00 \pm 0.00$ & $13.20 \pm 0.63$ & $\phantom{00}15.57 \pm \phantom{0}1.68$  \\
    Hepatitis & $19.77 \pm 7.16$ & $\phantom{0}4.30 \pm 0.95$ & \textit{no results} & \textit{no results} \\
    Lymphography & $14.37 \pm 7.31$ &  $\phantom{0}9.40 \pm 0.70$ & $11.20 \pm 2.10$  & $\phantom{0}172.30 \pm 48.32$ \\
    Mammographic & $20.43 \pm 4.03$ &  $\phantom{0}7.00 \pm 0.00$ & $8.00 \pm 0.00$ & $1547.13 \pm 76.04$ \\
    Mutagenesis & $\phantom{0}3.00 \pm 0.00$ &  $\phantom{0}3.00 \pm 0.00$ & \textit{timeout} & $\phantom{000}3.00 \pm \phantom{0}0.00$ \\
    NCTRER & $\phantom{0}3.00 \pm 0.00$ &  $\phantom{0}3.70 \pm 1.16$ & $7.00 \pm 0.00$ & $\phantom{000}3.00 \pm \phantom{0}0.00$ \\
    Premier League & $\phantom{0}7.00 \pm 0.00$ & $\phantom{0}9.00 \pm 0.00$ & $5.00 \pm 0.00$ & $\phantom{0}\phantom{0}20.03\pm \phantom{0}3.75$ \\
    Pyrimidine & $11.40 \pm 1.61$  &  $10.60 \pm 1.26$ & $6.60 \pm 1.26$ & $\phantom{00}19.67 \pm \phantom{0}4.49$  \\
    \bottomrule
  \end{tabular}
\end{table}

Table~\ref{tab:evaluation-results} evaluates our EvoLearner approach on the benchmarking datasets and compares it with state-of-the-art baselines. Our approach outperforms each of the baselines on at least~7 out of~9 datasets in terms of accuracy and $F_1$-measure. In each of the 8~pairwise comparisons of EvoLearner with a baseline according to $F_1$-measure or accuracy, we can reject the null hypothesis that both approaches are equally good with $p < 0.05$ (according to the non-parametric Wilcoxon signed-rank test). Looking at the lengths of learned concepts, we find that our approach generates much shorter concepts than SPaCEL and slightly longer concepts than CELOE and OCEL on most datasets. A manual inspection revealed that \mbox{SPaCEL} generates overly verbose expressions with many redundant and unnecessary parts, whereas CELOE's and OCEL's concepts are often incomplete. In the following, we show example solutions found by EvoLearner for the \textit{Premier League} and \textit{Carcinogenesis} learning problems. In case of \textit{Premier League}, the objective is to find descriptions of soccer goalkeepers based on their statistics; for example, the concept of a goalkeeper can be characterized by the number of shots on the target conceded. In case of \textit{Carcinogenesis}, molecules causing cancer can be characterized by certain chemical structures. Another example is the solution~\eqref{eq:expected-solution} found by EvoLearner for the \texttt{Uncle} problem.

\medskip

\noindent
\emph{Premier League:}\quad\\
$\texttt{Player} \sqcap \exists \texttt{has\_action.}(\texttt{shots\_on\_target\_conceded} \geq \texttt{2})$

\medskip

\noindent
\emph{Carcinogenesis:}\quad\\
$(\texttt{drosophila\_slrl} = \texttt{true}) \ \sqcup \ (\texttt{amesTestPositive} = \texttt{true}) \ \sqcup$
$(\geq 4 \ \texttt{hasStructure.Halide}) \ \sqcup$
$(\texttt{chromaberr} = \texttt{false})$

\begin{table}[tb]
  \centering
  \caption{$\boldsymbol{F_1}$-measure of 10-fold cross-validation of different configurations of EvoLearner. Note that the datasets \textit{Family} and \textit{Lymphography} do not contain data properties.}%
  \label{tab:ablation-analysis}
  \setlength{\tabcolsep}{2pt}
  \renewcommand{\arraystretch}{1.08}
  \footnotesize
  \begin{tabular}{@{}lcccc@{}}
    \toprule
     & \textbf{EvoLearner} & \textbf{Without} & \textbf{Without} & \textbf{Without} \\
    \textbf{Learning Problem} & \textbf{(ours)} & \textbf{Rand.\ Walk Init.} & \textbf{Data Properties} & \textbf{Both} \\
    \midrule
    Carcinogenesis &  \textbf{0.70} $\boldsymbol{\pm}$ \textbf{0.12} & 0.60 $\pm$ 0.21 & 0.63 $\pm$ 0.13 & 0.62 $\pm$ 0.13 \\
    
    Family & \textbf{1.00} $\boldsymbol{\pm}$ \textbf{0.01} & 0.87 $\pm$ 0.13 & --- & 0.86 $\pm$ 0.14\\
    
    Hepatitis & \textbf{0.79} $\boldsymbol{\pm}$ \textbf{0.08} & 0.67 $\pm$ 0.15 & 0.46 $\pm$ 0.14 & 0.47 $\pm$ 0.13\\
    
    Lymphography & \textbf{0.84} $\boldsymbol{\pm}$ \textbf{0.09} & 0.83 $\pm$ 0.11 & --- & 0.83 $\pm$ 0.09\\
    
    Mammographic & \textbf{0.81} $\boldsymbol{\pm}$ \textbf{0.06} & 0.78 $\pm$ 0.08 & 0.77 $\pm$ 0.07 & 0.75 $\pm$ 0.06\\
    
    Mutagenesis & \textbf{1.00} $\boldsymbol{\pm}$ \textbf{0.00} & \textbf{1.00} $\boldsymbol{\pm}$ \textbf{0.00} & 0.44 $\pm$ 0.48 & 0.50 $\pm$ 0.51\\
    
    NCTRER & \textbf{1.00} $\boldsymbol{\pm}$ \textbf{0.00} & \textbf{1.00} $\boldsymbol{\pm}$ \textbf{0.00} & 0.74 $\pm$ 0.05 & 0.75 $\pm$ 0.05\\
    
    Premier League & \textbf{1.00} $\boldsymbol{\pm}$ \textbf{0.00} & 0.98 $\pm$ 0.04 & 0.50 $\pm$ 0.23 & 0.50 $\pm$ 0.22\\
    
    Pyrimidine & \textbf{0.91} $\boldsymbol{\pm}$ \textbf{0.14} & 0.83 $\pm$ 0.22 & 0.67 $\pm$ 0.00 & 0.67 $\pm$ 0.00\\
    \bottomrule
  \end{tabular}
\end{table}

\subsection{Initialization and Data Properties}%
\label{subsec:evaluation-ablation-analysis}

To investigate the cause of EvoLearner's high performance, we perform an ablation study and investigate the following variants of EvoLearner: (1)~without random walk initialization, (2)~without data properties, and (3)~without both. Instead of our random walk initialization, we employ the \textsc{rampedHalfHalf} initialization method with a maximal depth of 6, which combines the \textsc{Full} and \textsc{Grow} methods and has been found to be one of the best default initialization methods~\citep{Koza1992GP}.

Table~\ref{tab:ablation-analysis} shows that both our novel initialization method and our support for data properties lead to an increase in performance on almost all datasets. For tackling the \textit{Hepatitis}, \textit{Mammographic}, \textit{Mutagenesis}, \textit{NCTRER} and \textit{Pyrimidine} learning problems, sufficient support for data properties and cardinality restrictions is crucial. Unsurprisingly, on the \textit{Lymphography} dataset, which does not contain any properties, neither our initialization nor support for data properties could boost performance. On \textit{Mutagenesis} and \textit{NCTRER}, we achieve an $F_1$-measure of $1.00$ with and without our initialization technique since the optimal solution only requires finding the suitable value for one data property.

\subsection{Variants of Biased Random Walks}%
\label{subsec:evaluation-walk-variants}

\begin{table}[tb]
    \centering
	\caption{$\boldsymbol{F_1}$-measure of 10-fold cross-validation of different variants of the random walk initialization. Note that \textit{Lymphography} only contains types but no properties.}%
	\label{tab:random_walk_init_comparison}
	\setlength{\tabcolsep}{4pt}
	\renewcommand{\arraystretch}{1.08}
	\footnotesize
	\begin{tabular}{@{}l@{\hskip -4pt}ccc|c@{}}
			\toprule
			& \textbf{Rand.\ Walk Init.} & \textbf{Without} & \textbf{Without} & \textbf{Without} \\
			
			\textbf{Learn.\ Probl.} & \textbf{(EvoLearner)} & \textbf{Paths} & \textbf{Types} & \textbf{Rand. Walk Init.} \\
			
			\midrule
			
			Carcinogenesis & \textbf{0.70} $\boldsymbol{\pm}$ \textbf{0.12} & $0.61 \pm 0.20$ & $0.65 \pm 0.17$ & 0.60 $\pm$ 0.21 \\
			
			Family & \textbf{1.00} $\boldsymbol{\pm}$ \textbf{0.01} & $0.86 \pm 0.14$ & $0.90 \pm 0.16$  & 0.87 $\pm$ 0.13 \\
			
			Hepatitis & \textbf{0.79} $\boldsymbol{\pm}$ \textbf{0.08} & $0.66 \pm 0.16$ & $0.76 \pm 0.13$  & 0.67 $\pm$ 0.15 \\
			
			Lymphography & \textbf{0.84} $\boldsymbol{\pm}$ \textbf{0.09} & \textbf{0.84} $\boldsymbol{\pm}$ \textbf{0.09} & --- & 0.83 $\pm$ 0.11 \\
			
			Mammographic & \textbf{0.81} $\boldsymbol{\pm}$ \textbf{0.06} & $0.78 \pm 0.08$ & $0.81 \pm 0.07$  & 0.78 $\pm$ 0.08 \\
			 
			Mutagenesis & \textbf{1.00} $\boldsymbol{\pm}$ \textbf{0.00} & \textbf{1.00} $\boldsymbol{\pm}$ \textbf{0.00} & \textbf{1.00} $\boldsymbol{\pm}$ \textbf{0.00} & \textbf{1.00} $\boldsymbol{\pm}$ \textbf{0.00} \\
			
			NCTRER & \textbf{1.00} $\boldsymbol{\pm}$ \textbf{0.00} & \textbf{1.00} $\boldsymbol{\pm}$ \textbf{0.00} & \textbf{1.00} $\boldsymbol{\pm}$ \textbf{0.00} & \textbf{1.00} $\boldsymbol{\pm}$ \textbf{0.00} \\
			
			Premier League & \textbf{1.00} $\boldsymbol{\pm}$ \textbf{0.00} & $0.98 \pm 0.05$ & \textbf{1.00} $\boldsymbol{\pm}$ \textbf{0.00} & 0.98 $\pm$ 0.04 \\
			
			Pyrimidine & \textbf{0.91} $\boldsymbol{\pm}$ \textbf{0.14} & $0.82 \pm 0.21$ & 0.87 $\pm$ 0.21 & $0.83 \pm 0.22$ \\
			
			\bottomrule
	\end{tabular}
\end{table}

We examined why the biased random walk method for initialization works so well by experimenting with different variants thereof: leaving out type or path information (see Section~\ref{subsec:initialization}). In
the first variant, we omitted Step (\ref{item1-rw}) from the random walk method, so we did not select a type during the initialization and only selected the paths afterward. For the second variant, we omitted Step (\ref{item2-rw}) and Step (\ref{item3-rw}) so the resulting concepts in the initial population only consisted of the selected type in Step (\ref{item1-rw}). Table~\ref{tab:random_walk_init_comparison}
shows that both type and path information are important to get the highest performance.  
However, path information was more important than type information, especially for \textit{Hepatitis} and \textit{Pyrimidine}. Note that the variant without types still outperforms the variant without random walks and the variant without paths is about as good.

\subsection{Dataset Characteristics}%
\label{subsec:evaluation-dataset-characteristics}

We observed that our random walk/bottom-up initialization starting at the ABox tends to be particularly good for datasets whose solutions require long class expressions and for datasets that contain many object properties in the TBox (Table~\ref{tab:evaluation-datasets} shows the number of object properties, Table~\ref{tab:evaluation-results} [bottom] the concept lengths, and Table~\ref{tab:evaluation-results} [top] the performance). This can be explained as follows: Top-down approaches randomly refining the top concept $\top$ have a hard time finding sensible concepts by chance if the class expressions are long and the TBox contains many object properties due to combinatorial explosion. On the other hand, our bottom-up initialization based on random walks only yields combinations of object properties that do appear in the ABox. As shown in Table~\ref{tab:ablation-analysis}, this effect is particularly pronounced on the datasets \textit{Carcinogenesis}, \textit{Family} and \textit{Hepatitis}. 

\subsection{Example of Emerging Concepts}%
\label{subsec:evaluation-example}

\begin{table*}[tb]
    \centering
    \caption{Emerging concepts for the \texttt{Uncle} learning problem.}%
    \label{tab:evaluation-example}
    \footnotesize
    \setlength{\tabcolsep}{2pt}
\begin{tabular}{@{}rlccccc@{}}
\toprule
  & & & \multicolumn{2}{c}{\textbf{Train}} & \multicolumn{2}{c}{\textbf{Test}} \\
  \cmidrule(lr){4-5} \cmidrule(lr){6-7}
  \textbf{Gen.} & \textbf{Concept} & \textbf{Len.} & \textbf{Acc.} & \textbf{F$_1$} & \textbf{Acc.} & \textbf{F$_1$} \\
  \midrule
  1 & $\texttt{Brother}$ $\sqcup$ $\exists \texttt{married.Sister}$ & 5 & 0.941 & 0.944 & 0.875 & 0.889 \\
  \midrule
  9 & $\texttt{Male} \sqcap ((\exists \texttt{married.}\exists\texttt{hasSibling.Daughter})$ $\sqcup $ $(\exists\texttt{hasSibling.}\exists\texttt{hasChild.}(  \leq 1 \  \texttt{hasParent.Male})))$ & 16 & 0.970 & 0.969 & 0.750 & 0.667\\
  \midrule
  11 & $\texttt{Male} \sqcap ((\exists \texttt{hasSibling.}\exists\texttt{hasChild.}(  \leq 1 \  \texttt{hasParent.Daughter}))$ $\sqcup $ $(\exists\texttt{married.}\exists\texttt{hasSibling.}\exists\texttt{hasChild}(  \leq 2 \  \texttt{hasParent.Male})))$ & 21 & 1.000 & 1.000 & 1.000 & 1.000 \\
  \midrule
  18 & $\texttt{Male} \sqcap ((\exists\texttt{hasSibling.}\exists \texttt{hasChild.Child})$ $\sqcup$  $(\exists\texttt{married.}\exists\texttt{hasSibling.} \exists\texttt{hasChild.Child}))$ & 15 & 1.000 & 1.000 & 1.000 & 1.000 \\
  \midrule
  23 & $\texttt{Male} \sqcap ((\exists\texttt{hasSibling.}\exists \texttt{hasChild.Child})$ $\sqcup$  $(\exists\texttt{married.}\exists\texttt{hasSibling.Parent}))$ & 13 & 1.000 & 1.000 & 1.000 & 1.000 \\
  \midrule  
  29 & $\texttt{Male} \sqcap ((\exists\texttt{hasSibling.Parent})$ $\sqcup$ $ (\exists\texttt{married.}\exists\texttt{hasSibling.Parent}))$ & 11 & 1.000 & 1.000 & 1.000 & 1.000 \\
\bottomrule
\end{tabular}
\end{table*}

Table~\ref{tab:evaluation-example} provides an example of emerging concepts from generation to generation on the \textit{Uncle} learning problem. From each generation's population, the best concept according to the fitness function is shown. It can be seen that up to generation 11, concepts become more accurate on the training set while also becoming longer. Afterwards, concepts maintain their good accuracy while becoming shorter. The best concept is reached after 29 generations. This pattern of growing and shrinking concepts is typical and we observed it for many datasets and learning problems.

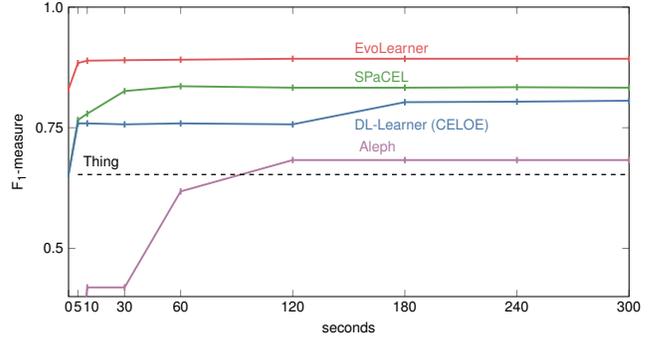
\begin{figure}[tb]
	\centering
	\tikzset{font=\fontfamily{phv}\selectfont}
	\begin{tikzpicture}[scale=0.6]
		\begin{axis}[
			legend pos=outer north east,
			legend cell align={left},
			xlabel=seconds,
			ylabel=F\textsubscript{1}-measure,
			xtick={0, 5, 10, 30, 60, 120, 180, 240, 300},
			xticklabels={0, 5, \ \ 10, 30, 60, 120, 180, 240, 300},
			ytick={0.25, 0.5, 0.75, 1.0},
			yticklabels={0.25, 0.5, 0.75, 1.0},
			xmin=0,
			xmax=300,
			ymin=0.4,
			ymax=1.0,
			width=14cm,
			height=8cm]
			\addplot [mark=|,  tab20darkred, very thick] coordinates {(0, 0.830) (5, 0.884) (10, 0.889) (30, 0.89) (60, 0.891) (120, 0.893) (180, 0.893) (240, 0.893) (300, 0.893)}
			node [pos=0.5, above right] {EvoLearner};
			\addplot [mark=|,  tab20darkgreen, very thick] coordinates {(0, 0.653)(5, 0.766) (10, 0.779) (30, 0.826) (60, 0.836) (120, 0.833) (180, 0.833) (240, 0.834) (300, 0.833)}
			node [pos=0.5, above right] {SPaCEL};
			\addplot [mark=|,  tab20darkblue, very thick] coordinates {(0, 0.653)(5, 0.759) (10, 0.759) (30, 0.757) (60, 0.759) (120, 0.757) (180, 0.803) (240, 0.804) (300, 0.806)}
			node [pos=0.5, below right] {DL-Learner (CELOE)};
			\addplot [mark=|,  tab20darkpurple, very thick] coordinates {(5, 0.248) (10, 0.419) (30, 0.419) (60, 0.618) (120, 0.683) (180, 0.683) (240, 0.683) (300, 0.683)}
			node [pos=0.5, above right] {Aleph};
			\addplot [mark=none,  thick, dashed] coordinates {(5, 0.653) (10, 0.653) (30, 0.653) (60, 0.653) (120, 0.653) (180, 0.653) (240, 0.653) (300, 0.653)}
			node [pos=0, above right] {Thing};
			Si    	\end{axis} 
	\end{tikzpicture}%
	\vspace{-2ex}%
	\caption{Average $\boldsymbol{F_1}$-measure depending on runtime over all datasets except \textit{Hepatitis} and \textit{Family} since not all approaches yielded results for them.%
	}%
	\label{fig:runtime-effectiveness}%
	\Description{Average F1-measure depending on runtime over all datasets except Hepatitis and Family since not all approaches yielded results for them}
	\vspace{-2ex}
\end{figure}

\subsection{Effectiveness Depending on Runtime}%
\label{subsec:effectiveness-over-runtime}

In Figure~\ref{fig:runtime-effectiveness}, we examine the relationship between performance and runtime. The time for loading the datasets is not taken into account. For EvoLearner, the time starts right before the calculation of the splits and the random-walk initialization; for SPaCEL and CELOE, time starts after the initialization of the search and before the start of the refinements; for Aleph, time starts immediately as per SML-Bench. After the random-walk initialization, EvoLearner achieves an $F_1$-measure of~0.83 averaged over all datasets. This number increases to~0.88 after the evolutionary algorithm has run for 5~seconds. Except for \textit{Carcinogenesis}, EvoLearner reaches 200 generations within the first 60 seconds, with further generations hardly increasing performance. A manual analysis revealed that for \textit{NCTRER} and \textit{Mutagenesis}, our initialization almost always generated an individual with the correct data property and split within the initial population. SPaCEL and CELOE initially predict \texttt{Thing}, yielding an $F_1$-measure of~0.65. SPaCEL finishes in under a second for most datasets, except for \textit{Carcinogenesis}, \textit{Lymphography} and \textit{Mammographic}. Therefore, performance hardly improves afterwards. CELOE struggles to improve on \textit{Carcinogenesis}, \textit{Lymphography}, \textit{Mammographic}, \textit{Mutagenesis} and \textit{NCTRER}. On \textit{Pyrimidine}, the performance of CELOE even decreases over time due to overfitting. The jump between 2 and 3 minutes is due to an increase in $F_1$-measure on \textit{Premier League}. Aleph does not provide partial solutions, and the jumps of average performance correspond to Aleph's solution to \textit{NCTRER} after about 10~seconds and Aleph's solution to \textit{Carcinogenesis} and \textit{Premier League} after about 40~seconds.

\section{Discussion}%
\label{sec:discussion}

\paragraph{Open-world vs.\ Closed-world Reasoning.} Closed-world reasoners~\citep{Kroetzsch2012Description} assume every fact not explicitly stated to be true to be actually false. Open-world reasoners keep unspecified information open and require a logical consequence to be true in all conceivable states of the world. In this work, we employ a closed-world reasoner for the following reasons: (1)~Our baseline systems employ closed-world reasoning. (2)~
Closed-world reasoners may be less surprising to new users%
\footnote{\url{https://spinrdf.org/shacl-and-owl.html}}
~\cite{Ruttenberg2005Experience}.
(3)~Closed-world reasoners are a lot faster, allowing the evaluation of many more concepts in the same amount of time. Our code can be adapted to employ an open-world reasoner, which is available as part of the Owlready2 library~\cite{Lamy2017OWL}.

\paragraph{Datatypes.}
Our current implementation supports Booleans as well as numbers including double values and integers. Extending this to dates is straightforward as dates can be sorted, and thresholds can be computed in the same way as for numbers. An extension to strings can be realized by treating them as atomic concepts.

\paragraph{Initialization of Baselines}
Our initialization strategy is specifically tailored towards evolutionary algorithms and our crossover and mutation operators. Applying our initialization strategy to our baselines would lead to a significantly smaller search space since concepts that are more general than our initial concepts would not be found by downward refinement operators. We leave it to future work to develop novel initialization strategies specifically tailored to inductive logic programming with refinement operators.

\section{Conclusion}%
\label{sec:conclusion}

In this paper, we introduce EvoLearner, an approach to learn description logic concepts with evolutionary algorithms. In particular, our contributions are twofold: (1)~a novel initialization method based on biased random walks, and (2)~support for data properties, which maximizes information gain. Our results show that EvoLearner significantly outperforms the state of the art, and our ablation study confirms that this can be attributed to our initialization and support for data properties. An analysis of performance over runtime indicates that we find equally good concepts faster than the state of the art, and our approach often keeps improving when other approaches are already stagnating.

In future work, we will explore ways to combine our approach with knowledge graph embeddings~\cite{Demir2021ConvolutionalComplex, Demir2021ConvolutionalHypercomplex} and reinforcement learning~\cite{Demir2021DRILL}, e.g., guiding the crossover and mutation in a promising direction. Moreover, we will further increase the expressiveness of our approach towards OWL 2 Full, i.e., $\mathcal{SROIQ(D)}$. Applications may include fact checking~\cite{Syed2019Unsupervised} and vandalism detection~\cite{Heindorf2016Vandalism, Heindorf2019Debiasing}.

\begin{acks}
This work has been supported by the German Federal Ministry for Economic Affairs and Energy (BMWi) within the project RAKI under the grant no 01MD19012B and by the German Federal Ministry of Education and Research (BMBF) within the project DAIKIRI under the grant no 01IS19085B. We are grateful to Pamela Heidi Douglas for editing the manuscript.
\end{acks}

\clearpage

{%
\raggedright
\bibliographystyle{ACM-Reference-Format}

\clearpage
}%
\section{Appendix}
In the following, we show the results of additional experiments we performed regarding initialization methods, mutation operators, and hyperparameters. Moreover, we provide pseudocode for our random walk initialization and our calculation of splitting values for data properties.

\subsection{Comparison of Initialization Methods}

In Table~\ref{tab:init_comparison}, we compare our random walk initialization to three widely used initialization methods~\citep{Koza1992GP}: \textsc{Grow}, \textsc{Full}, and \textsc{RampedHalfHalf}. \textsc{Grow} generates trees where leaves can have different depths, \textsc{Full} generates trees where all leaves have the same depth, and \textsc{RampedHalfHalf} generates half of a population's individuals with the \textsc{Grow} or \textsc{Full} method. We observe that the performance of the three methods is similar, with \textsc{Full} having a slight edge, while our random walk initialization performs significantly better. We hypothesize that this can be attributed to the fact that our generated class expressions, i.e., combinations of atomic concepts and operators, do actually appear in the instance data while the other three methods generate random/artificial combinations unrelated to the instance data. Manual spot checks confirmed our hypothesis.

\subsection{Comparison of Mutation Operators}

Table~\ref{tab:mutation} compares the performance of different mutation operators which are part of the DEAP framework~\citep{Fortin2012DEAP}:  \texttt{mut\-Uniform}, \texttt{mut\-Shrink}, \texttt{mut\-Node\-Re\-place\-ment} and \texttt{mut\-Insert}.
\texttt{mut\-Uniform} uniformly randomly selects a node in the tree and replaces it by a randomly generated tree as described in Section~\ref{subsec:evolutionary-algorithm}; \texttt{mut\-Shrink} randomly selects two nodes in the tree which are on a path from the root to a leaf and replaces the subtree induced by the higher node by the subtree induced by the lower node. \texttt{mut\-Node\-Re\-place\-ment} randomly replaces a single node in the tree by a single random node; \texttt{mut\-Insert} randomly selects an inner node, replaces it with a random inner node, and adds the original subtree induced by the node as a child.

While the differences are small on most datasets, \texttt{mut\-Uniform} performs much better than other approaches on \textit{Hepatitis}. We attribute this to the fact that \texttt{mut\-Uniform} introduces new constructs, like universal restrictions, negations, and cardinality restrictions, which \texttt{mut\-Shrink} does not.
Similarly, \texttt{mut\-Node\-Re\-place\-ment} and \texttt{mut\-Insert} can only introduce them to a limited extent by replacing or adding single nodes. 
These constructs seem to be particularly important for the Hepatitis dataset where almost all the good solutions contain them. 
In case of \textit{Lymphography}, \texttt{mut\-Uniform} learns long class expressions with training scores exceeding $0.9$ (not shown in tables) which cannot be matched on test data (0.84 in Table~\ref{tab:mutation}), hinting at slight overfitting.

\begin{table}[tbp]
    \centering
	\caption{Different initialization methods in terms of $\boldsymbol{F_1}$-measure for 10-fold cross-validation.}%
	\label{tab:init_comparison}
	\setlength{\tabcolsep}{5.9pt}
	\footnotesize
	\begin{tabular}{@{}lcccc@{}}
			\toprule
			\textbf{Learning Problem} & \textbf{Rand.\ Walk} & \textbf{\textsc{Grow}} & \textbf{\textsc{Full}} & \textbf{\textsc{Ramped}} \\
			\midrule
			Carcinogenesis & \textbf{0.70} $\boldsymbol{\pm}$ \textbf{0.12} & $0.61 \pm 0.20$ & $0.63 \pm 0.19$ & $0.60 \pm 0.21$ \\
			
			Family & \textbf{1.00} $\boldsymbol{\pm}$ \textbf{0.01} & $0.86 \pm 0.14$ & $0.87 \pm 0.14$ & $0.87 \pm 0.13$\\
			
			Hepatitis & \textbf{0.79} $\boldsymbol{\pm}$ \textbf{0.08} & $0.63 \pm 0.20$ & $0.76 \pm 0.14$ & $0.67 \pm 0.15$\\
			
			Lymphography & \textbf{0.84} $\boldsymbol{\pm}$ \textbf{0.09} & $0.83 \pm 0.08$ & \textbf{0.84} $\boldsymbol{\pm}$ \textbf{0.09} &  $0.83 \pm 0.11$\\
			
			Mammographic & \textbf{0.81} $\boldsymbol{\pm}$ \textbf{0.06} & $0.78 \pm 0.08$ & $0.78 \pm 0.08$ & $0.78 \pm 0.08$\\
			 
			Mutagenesis & \textbf{1.00} $\boldsymbol{\pm}$ \textbf{0.00} & \textbf{1.00} $\boldsymbol{\pm}$ \textbf{0.00} & \textbf{1.00} $\boldsymbol{\pm}$ \textbf{0.00} & \textbf{1.00} $\boldsymbol{\pm}$ \textbf{0.00}\\
			
			NCTRER & \textbf{1.00} $\boldsymbol{\pm}$ \textbf{0.00} & \textbf{1.00} $\boldsymbol{\pm}$ \textbf{0.00} & \textbf{1.00} $\boldsymbol{\pm}$ \textbf{0.00} & \textbf{1.00} $\boldsymbol{\pm}$ \textbf{0.00}\\
			
			Premier League & \textbf{1.00} $\boldsymbol{\pm}$ \textbf{0.00} & 0.99 $\pm$ 0.04  & 0.98 $\pm$ 0.05 & 0.98 $\pm$ 0.04 \\
			
			Pyrimidine & \textbf{0.91} $\boldsymbol{\pm}$ \textbf{0.14} & $0.84 \pm 0.22$ & $0.84 \pm 0.22$ & $0.83 \pm 0.22$\\
			\bottomrule
	\end{tabular}
\end{table}
\begin{table}[tbp]
    \centering
	\caption{Different mutation operators in terms of $\boldsymbol{F_1}$-mea\-sure for 10-fold cross-validation.}%
	\label{tab:mutation}
	\setlength{\tabcolsep}{6pt}
	\footnotesize
	\begin{tabular}{@{}lcccc@{}}
			\toprule
			\textbf{Learning Problem} & \textbf{\texttt{mutUniform}} & \textbf{\texttt{mutShrink}} &
			\textbf{\texttt{mutN.Repl.}} & \textbf{\texttt{mutInsert}} \\
			\midrule
			Carcinogenesis & \textbf{0.70} $\boldsymbol{\pm}$ \textbf{0.12} & $0.65 \pm 0.17$& $0.68 \pm 0.15$& $0.66 \pm 0.14$ \\
			 
			Family & \textbf{1.00} $\boldsymbol{\pm}$ \textbf{0.01} & $0.96 \pm 0.06$ & $0.99 \pm 0.02$& $0.97 \pm 0.06$\\
			
			Hepatitis & \textbf{0.79} $\boldsymbol{\pm}$ \textbf{0.08} & $0.40 \pm 0.21$& $0.53 \pm 0.16$& $0.49 \pm 0.23$\\
			
			Lymphography & \textbf{0.84} $\boldsymbol{\pm}$ \textbf{0.09} & $0.81 \pm 0.09$ & $0.81 \pm 0.09$ & $0.83 \pm 0.09$\\
			
			Mammographic & \textbf{0.81} $\boldsymbol{\pm}$ \textbf{0.06} & $0.80 \pm 0.07$& $0.81 \pm 0.07$& $0.80 \pm 0.06$\\
			 
			Mutagenesis & \textbf{1.00} $\boldsymbol{\pm}$ \textbf{0.00} & $0.97 \pm 0.11$ & \textbf{1.00} $\boldsymbol{\pm}$ \textbf{0.00} & $0.97 \pm 0.11$ \\
			
			NCTRER & \textbf{1.00} $\boldsymbol{\pm}$ \textbf{0.00} & $0.99 \pm 0.02$ & \textbf{1.00} $\boldsymbol{\pm}$ \textbf{0.00} & $0.99 \pm 0.02$\\
			
			Premier League & \textbf{1.00} $\boldsymbol{\pm}$ \textbf{0.00}& $0.98 \pm 0.05$ & $0.99 \pm 0.03$ & $0.98 \pm 0.06$ \\
			
			Pyrimidine & 0.91 $\pm$ 0.14 & $0.91 \pm 0.15$& $0.92 \pm 0.13$ & \textbf{0.93} $\boldsymbol{\pm}$ \textbf{0.14}\\
			
			\bottomrule
	\end{tabular}
\end{table}
\begin{table}[tbp]
    \centering
	\caption{Different settings for $\boldsymbol{maxT}$ in terms of $\boldsymbol{F_1}$-mea\-sure for 10-fold cross-validation.}%
	\label{tab:maxt}
	\setlength{\tabcolsep}{6pt}
	\footnotesize
	\begin{tabular}{@{}lcccc@{}}
			\toprule
			\textbf{Learning Problem} & \textbf{maxT = 1} & \textbf{maxT = 2} &
			\textbf{maxT = 4} & \textbf{maxT = 6} \\
			\midrule
			Carcinogenesis  & $0.64 \pm 0.18$& \textbf{0.70} $\boldsymbol{\pm}$ \textbf{0.12} & $0.67 \pm 0.14$ & $0.68 \pm 0.14$  \\
			 
			Family & $0.93 \pm 0.08$ & 1.00 $\pm$ 0.01 & $0.98 \pm 0.05$ & \textbf{1.00} $\boldsymbol{\pm}$ \textbf{0.00}\\
			
			Hepatitis & $0.71 \pm 0.15$ & 0.79 $\pm$ 0.08 & $0.79 \pm 0.12$ & \textbf{0.82} $\boldsymbol{\pm}$ \textbf{0.10}\\
			
			Lymphography & \textbf{0.84} $\boldsymbol{\pm}$ \textbf{0.09} & \textbf{0.84} $\boldsymbol{\pm}$ \textbf{0.09} & \textbf{0.84} $\boldsymbol{\pm}$ \textbf{0.09} & \textbf{0.84} $\boldsymbol{\pm}$ \textbf{0.09}\\
			
			Mammographic & $0.79 \pm 0.08$ & $0.81 \pm 0.06$ & \textbf{0.82} $\boldsymbol{\pm}$ \textbf{0.05} & $0.81 \pm 0.07$\\
			 
			Mutagenesis & \textbf{1.00} $\boldsymbol{\pm}$ \textbf{0.00} & \textbf{1.00} $\boldsymbol{\pm}$ \textbf{0.00} & \textbf{1.00} $\boldsymbol{\pm}$ \textbf{0.00} & \textbf{1.00} $\boldsymbol{\pm}$ \textbf{0.00}\\
			
			NCTRER & \textbf{1.00} $\boldsymbol{\pm}$ \textbf{0.00} & \textbf{1.00} $\boldsymbol{\pm}$ \textbf{0.00} & \textbf{1.00} $\boldsymbol{\pm}$ \textbf{0.00} & \textbf{1.00} $\boldsymbol{\pm}$ \textbf{0.00}\\
			
			Premier League & \textbf{1.00} $\boldsymbol{\pm}$ \textbf{0.00} & \textbf{1.00} $\boldsymbol{\pm}$ \textbf{0.00} & $1.00 \pm 0.02$ & \textbf{1.00} $\boldsymbol{\pm}$ \textbf{0.00} \\
			
			Pyrimidine & $0.85 \pm 0.21$ & \textbf{0.91} $\boldsymbol{\pm}$ \textbf{0.14} & $0.91 \pm 0.15$ & $0.87 \pm 0.17$\\
			
			\bottomrule
	\end{tabular}
\end{table}

\subsection{Comparison of maxT Settings}

The hyperparameter $maxT$ considers the number of outgoing triples of positive examples (c.f., Section~\ref{subsec:initialization} and Section~\ref{subsec:pseudocode}). Table~\ref{tab:maxt} evaluates different values of the $maxT$ parameter. It shows that a value of two is needed since setting $maxT:=1$ yielded worse $F_1$-measures. However, increasing the parameter beyond $maxT:=2$ did not significantly increase the $F_1$-measure and resulted in longer solutions. 

\subsection{Pseudocode}
\label{subsec:pseudocode}

Algorithm~\ref{alg:initialization} depicts the main loop of our random walk initialization. 
The \texttt{PopulationFromExamples} method takes a set of positive examples $E^+$ to generate the initial population. Therefore, it randomly selects positive examples $e^+ \in E^+$ and calls \texttt{ConceptFromExample}, which performs a biased random walk on $e^+$ to generate a concept. 
Algorithm~\ref{alg:initialization-property} turns a role/object pair $(r, o)$ into an existential restriction in case of an object property. In case of a data property, it generates a Boolean value restriction or a min/max numeric value restriction. 
Algorithm~\ref{alg:data_properties} shows the pseudocode for the calculation of splitting values for data properties. If not otherwise specified, sampling in the algorithms is performed uniformly at random.

\clearpage
\begin{algorithm}
	\DontPrintSemicolon
	\caption{Generates the initial population by creating concepts from positives examples via biased random walks.}%
	\label{alg:initialization}
	\textbf{Input:} Knowledge base $\mathcal{K}$, Positive examples $E^{+}$, Max.\ adjacent triples $\mathit{maxT}$, Set of $splits$ for each data property\\
    	\textbf{Output:} Set of concepts $population$ \\
    	\SetKwFunction{FMain}{PopulationFromExamples}
    	\SetKwProg{Pn}{Function}{:}{\KwRet $population$}
    	\Pn{\FMain{$\mathcal{K}$, $E^+$, $maxT$, $splits$}}{%
    		Compute the frequencies $\mathit{ct}[C]$ of the types (=atomic concepts) $C$ of the positive examples $E^{+}$\;
    		$population = []$\;
    		\While{$len(population) <$ \textsc{PopulationSize}}{%
    		Uniformly randomly pick $e^{+} \in E^{+}$\;
    		$population.append(  $ \texttt{ConceptFromExample}${(\mathcal{K}, e^{+}, \mathit{ct}, \mathit{maxT}, \mathit{splits})})$\;		
    		}
    	}
    \BlankLine
    \BlankLine
	\textbf{Input:} Knowledge base $\mathcal{K}$, Positive example $e^+$, Type frequencies $\mathit{ct}$, Max.\ adjacent triples $\mathit{maxT}$, Set of $splits$ for each data property \\
	\textbf{Output:} Concept $conc$ \\
	\SetKwFunction{F}{ConceptFromExample}
	\SetKwProg{Fn}{Function}{:}{\KwRet $conc$}
	\Fn{\F{$\mathcal{K}$, $e^+$, $\mathit{ct}$, $\mathit{maxT}$, $\mathit{splits}$}}{%
		Sample an atomic concept $C$ from $\{C \ |\ \mathcal{K} \models C(e^+)\}$ weighted by the frequencies $ct$ \tcp*[r]{Step (\ref{item1-rw})}
		$conc = C$\;
		$R = \{r \ | \ \mathcal{K} \models r(e^+, \cdot)\}$ \tcp*[r]{Step (\ref{item2-rw})}
		Sample up to $maxT$ $roles$ from $R$
		without replacement\;
		$role\_obj = \emptyset$\;
		\For{\textbf{each} $r \in roles$}{%
			Sample $o$ from $\{o \ |\ \mathcal{K} \models r(e^+, o) \}$
			\;
			$role\_obj = role\_obj \cup \{(r, o) \}$\;
		}
		\If{$|role\_obj|<maxT$}{%
		    $num = maxT - |role\_obj|$ \;
			Sample up to $num$ $new\_role\_obj$ from $\{(r,o) \ | \ \mathcal{K} \models r(e^+, o) \wedge r \in R  \}$
			without replacement\;
			$role\_obj = role\_obj \cup new\_role\_obj$\;
		}
		\For($\qquad\qquad \; \; \; \ $\tcp*[h]{Step (\ref{item3-rw})}){\textbf{each} $(r, o) \in role\_obj$}{%
			Sample an operator $op$ from $\{\sqcup, \sqcap \}$
			\;
			\eIf{ $type(r) = DataProperty $ \textbf{or} $\{s \ | \ \mathcal{K} \models s(o, \cdot)\} = \emptyset $ \textbf{or} $random() < 0.5$ }{%
				    $conc = conc$ op \texttt{RoleObjToConc}($\mathcal{K}$, $(r, o)$, $splits$)\; 
			    } 
			 {%
			    Sample a role $s$ from $\{s \ | \ \mathcal{K} \models s(o, \cdot)\}$
			    \;
				Sample $v$ from $\{(v \ | \ \mathcal{K} \models s(o, v) \wedge v \neq e^+\}$
				\;
				$conc = conc \ op \ \exists r$.\texttt{RoleObjToConc}($\mathcal{K}$, $(s, v)$, $splits$)\; 
			 }
		}
	}
\end{algorithm}
\begin{algorithm}
	\DontPrintSemicolon
	\caption{Turns the role/object pair into a concept.}%
	\label{alg:initialization-property}
    \textbf{Input:} Knowledge base $\mathcal{K}$, Role/Object Pair $(r, o)$, Set of $splits$ for each data property\\
    	\textbf{Output:} Concept $conc$ \\
    	\SetKwFunction{FMain}{RoleObjToConc}
    	\SetKwProg{Pn}{Function}{:}{\KwRet $conc$}
    	\Pn{\FMain{$\mathcal{K}$, $(r, o)$, $splits$}}{%
    	
    	\uIf{$type(o) = Individual$ }{%
    	        Sample an atomic concept $D$ from $\{D \ | \ \mathcal{K} \models D(o) \}$, if the set is empty take Thing\;
				$conc = (\exists r.D)$\; 
			    } 
			 \uElseIf{$type(o) = Boolean$}{%
			 	$conc = (r = o)$\;   
			 }
			\uElseIf{$type(o) = numeric$}{%
			        Find the split $v \in splits[r]$ that is closest to o\;
			        \eIf{$o \geq v$ } {%
			            $conc = (r \geq v)$\;   
			        }{%
			            $conc = (r \leq v)$\;  
			        }
			 }
    	}
\end{algorithm}
\newpage
\begin{algorithm}
	\DontPrintSemicolon
	\caption{Calculation of splitting values for data properties.}%
	\label{alg:data_properties}

	\textbf{Input:} Knowledge base $\mathcal{K}$, Data properties $P$, Number of splits $k$, Positive examples $E^{+}$, Negative examples $E^{-}$ \\
	\textbf{Output:} Set of $splits$ for each data property $d \in P$ \\
	\SetKwFunction{FMain}{CalculateSplits}
	\SetKwProg{Pn}{Function}{:}{\KwRet $splits$}
	\Pn{\FMain{$\mathcal{K}$, $P$, $k$, $E^+$, $E^-$}}{%
		For each $d \in P$ set $splits[d] = \emptyset$\;
		$current\_sets = \{E^+ \cup E^-\}$\;
		\While{$ P \neq \emptyset $ \textbf{and} $current\_sets \neq \emptyset$}{%
			$next\_level = \emptyset$\;
			\For{\textbf{each} $d \in P$}{%
					\For{\textbf{each} $E \in current\_sets$}{%
						$V^E_d = \{v \ | \ \mathcal{K} \models d(e,v) \wedge e \in E \}$\;
						Sort $V^E_d$ in ascending order\;
						$\bar{V}^E_d = \{(v_i  + v_{i+1}) /2 \ | \ i \in \{0,1,\ldots,|V^E_d| - 2\}\}$\;
						Find the best split $\bar{v} \in \bar{V}^E_d$ on $E$ depending on the information gain\;
						$splits[d] = splits[d] \cup \{ \bar{v} \}$\;
						Let $E^L$ and $E^R$ be the sets produced by splitting $E$ on $\bar{v}$\;
						$next\_level = next\_level \cup \{E^L, E^R\}$\;
						\If{$|splits[d]| \geq k$}{%
						    $P.remove(d)$\;
						    $break$\;
					    }
				}
			}
			$current\_sets = next\_level$\;
			Sort $current\_sets$ in descending order on the remaining entropy of the sets and remove sets with $0$ entropy\;
		}
	}
\end{algorithm}
\end{document}